\documentclass{article} 
\usepackage{iclr2016_workshop,times}
\usepackage{hyperref}
\usepackage{url}
\usepackage{tabularx}
\usepackage{hhline}
\usepackage{multirow}
\usepackage{graphicx} 
\usepackage{algorithm}
\usepackage{algorithmic}
\usepackage{amsthm}
\usepackage{amsmath}
\usepackage{amsfonts}
\usepackage{bbm}
\usepackage{titlesec}
\usepackage{caption}
\usepackage{subcaption}
\usepackage{natbib}
\usepackage{csquotes}
\usepackage{appendix}
\usepackage{subcaption}

\title{Stuck in a What?  \\ 
Adventures in Weight Space}

\author{Zachary C. Lipton \thanks{\url{http://zacklipton.com}} \\
Department of Computer Science \& Engineering\\
University of California, San Diego\\
La Jolla, CA 92093 , USA \\
\texttt{zlipton@cs.ucsd.edu} \\
}

\begin{document}

\maketitle

\begin{abstract}
Deep learning researchers commonly suggest 
that converged models are stuck in local minima.
More recently, some researchers observed 
that under reasonable assumptions, 
the vast majority of critical points are saddle points, not true minima.
Both descriptions suggest that weights converge around  a point in weight space, be it a local optima or merely a critical point.
However, it's possible that neither interpretation is accurate.
As neural networks are typically over-complete,
it's easy to show the existence of  vast  continuous regions through weight space with equal loss.
In this paper, we build on recent work empirically characterizing the error surfaces of neural networks.
We analyze training paths through weight space,
presenting evidence that apparent convergence of loss
does not correspond to weights arriving at critical points, 
but instead to large movements through flat regions of weight space.
While it's trivial to show that neural network error surfaces are globally non-convex, 
we show that error surfaces are also locally non-convex, even after breaking symmetry with a random initialization and also after partial training.
\end{abstract}

\section{Introduction}

In the worst case, solving for the optimal weights in a neural network is an NP-Hard problem.
Further, the error surfaces of neural networks are highly non-convex,
presenting seemingly formidable obstacles to learning by gradient descent.
And yet practitioners train deep neural networks everyday by stochastic gradient descent,
achieving state of the art results on a broad range of tasks.
In fact, for many problems, this method easily achieves zero loss on the training set.
Thus, while optimization presents a tremendous problem in theory,
one might argue that in practice, 
regularization is the greater concern.  

This disparity, between the apparent hopelessness of the optimization problem
and the \emph{de facto} ease of training 
has spurred several researchers to attempt 
both theoretically and empirically to characterize the error surfaces of deep neural networks.
Notably \citet{goodfellow2014qualitatively},
plotted loss along straight lines through weight space,
between two converged models, 
showing monotonic increases and then decreases in loss.
One might ask: 
\emph{Once symmetry is broken, 
is the problem convex?} 
Of course, the gradient at any point along this line doesn't necessarily point directly towards the nearest minimum. 
\citet{dauphin2014identifying} presented a case based on both empirical study and results from statistical physics suggesting that the ratio of saddle points to local minima
on a neural network's loss surface 
grows exponentially in the number of parameters. 
\citet{janzamin2015beating} presented a theoretical study, showing that under reasonable conditions on the data, the optimization problem can be made and solved via tensor decomposition.

\subsection{Contributions}
In this paper, we conduct preliminary experiments, training a standard three layer convolutional neural network with 819557 parameters on the MNIST dataset \citep{lecun1998mnist}, using dropout regularization and $\ell_2^2$ weight decay. 
We analyze the paths through weight space taken over the course of gradient descent, 
presenting the following findings:
\begin{itemize}
\item Weights do not converge to critical points,
instead traveling large (euclidean) distances 
through flat basins in weight space.
\item  While a straight line in weight-space from initialization to solution 
may correspond to monotonically decreasing loss,
the path actually taken by gradient descent seems far from straight.
\item A small number of principal components explains most of the variance along a training trajectory.
\item Even once symmetry is broken, neural network error surfaces are neither convex nor quasi-convex but continue to diverge towards many different low error basins. Starting from the same initialization, but then feeding each network examples in shuffled order 
is sufficient to diverge each network along a different path. 
This suggests that the error surface is not only globally non-convex, but also locally non-convex even for a partially trained net.
\item All pairs of solutions after a fixed number of epochs  appear to be roughly the same euclidean distance from the origin and from each other. This is true even with identical initializations, and pretraining before cloning.

\end{itemize}

\section{Experiments}
Rather than plotting straight lines through weight space like \citet{goodfellow2014qualitatively},
we investigate the paths through weight space taken as models are trained.
We analyze these trajectories qualitatively 
by visualizing them via 2D PCA, 
and quantitatively by analyzing the variance 
explained by the largest principal components.

\begin{figure}[t]
\centering
\begin{subfigure}{.32\textwidth}
  \centering
  \includegraphics[width=1\linewidth]{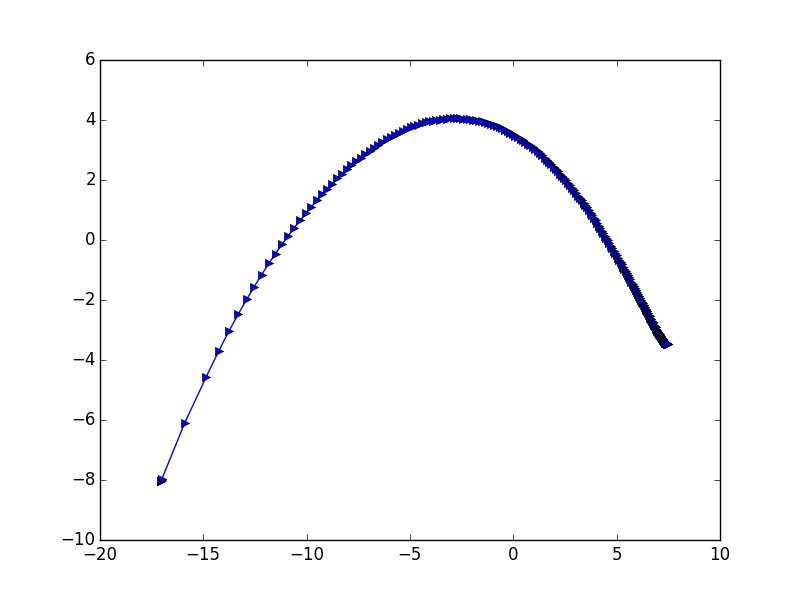}
  \label{fig:200epochs}
\end{subfigure}%
\begin{subfigure}{.32\textwidth}
  \centering
  \includegraphics[width=1\linewidth]{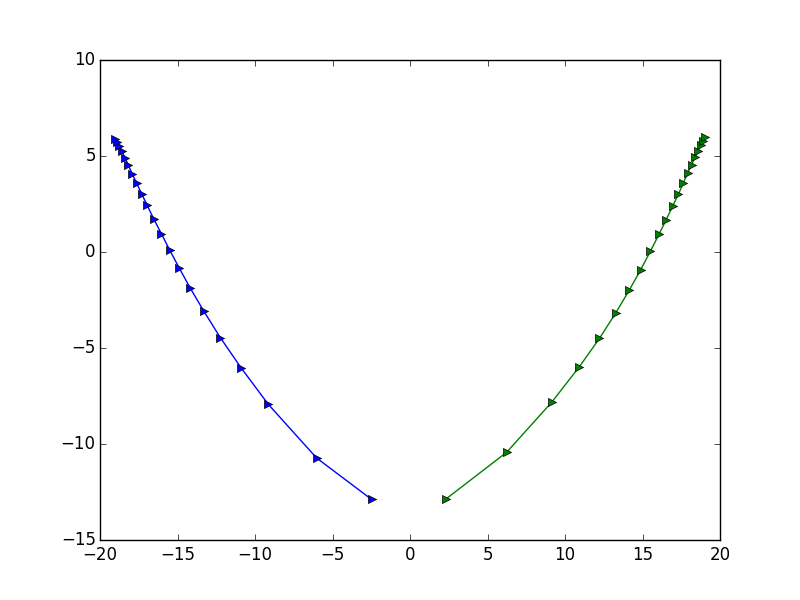}
\end{subfigure}
\begin{subfigure}{.32\textwidth}
  \centering
  \includegraphics[width=1\linewidth]{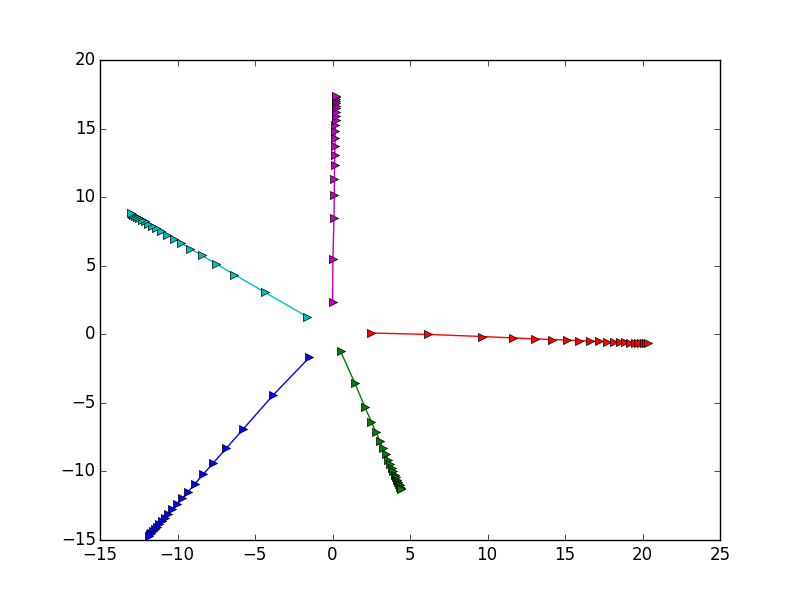}
\end{subfigure}
\caption{2D PCA of 1, 2, and 5 paths through weight-space, each from a different random initialization. Pairwise euclidean distances between all solutions are all roughly equivalent.}
\label{fig:pcaplots}
\end{figure}

We train a single model for 200 epochs, capturing its full parameters after each epoch. We plot this trajectory with a 2D PCA showing the high degree of non-linearity in the learned path. We then train 5 models for 200 epochs each, starting starting each model from different random initializations, and capturing their parameters every ten epochs.

Next, we repeat this same experiment but with the same initialization. We train each of the $5$ networks on a different shuffle of the data from the same starting point in weight space.

Finally, we train one model for 10 epochs. Then we clone it 5 times. Each clone is trained from these partially learned starting weights but with a different random shuffle.

\section{Results}
To visually demonstrate that the paths taken though weight space are highly nonlinear, we plot a $2D$ PCA of a single $200$ epoch trajectory (\autoref{fig:pcaplots}). The first two principal components explain $81.39\%$ of the variance. The top $10$ principal components explain $95.63\%$ of the variance. Speculatively, it seems that the low dimension of the trajectories, together with the smoothness of the curves might be useful properties for projecting where to look next.

\begin{figure}[ht]
\centering
\begin{subfigure}{.32\textwidth}
  \centering
  \includegraphics[width=1\linewidth]{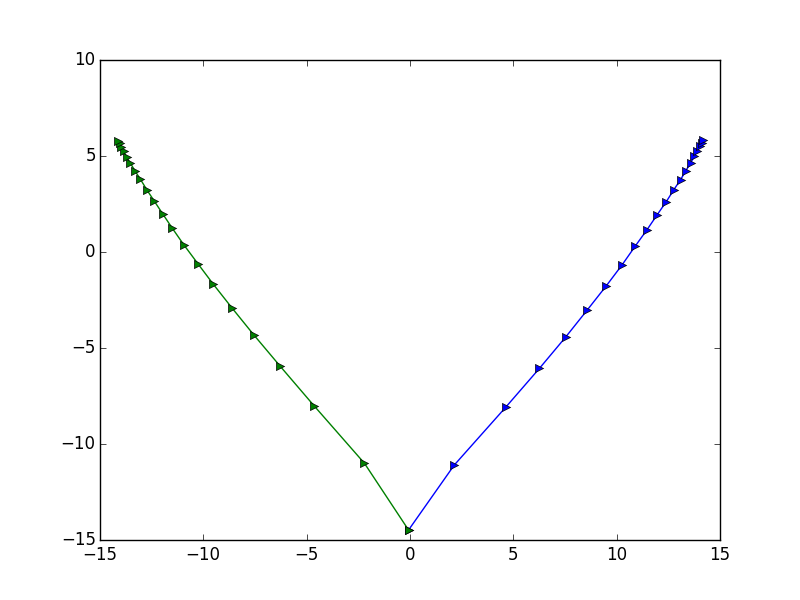}
\end{subfigure}
\begin{subfigure}{.32\textwidth}
  \centering
  \includegraphics[width=1\linewidth]{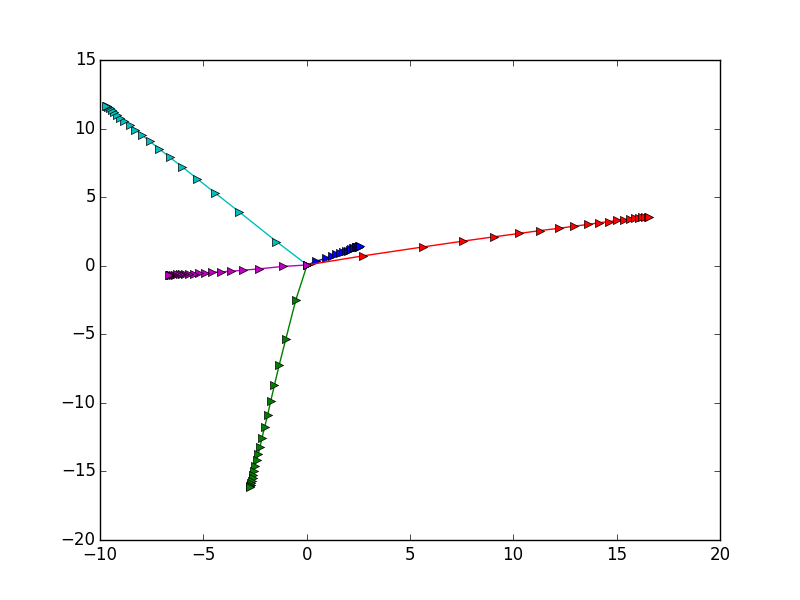}
\end{subfigure}
\caption{Paths through weight-space, each from an identical random initialization but with a different shuffle of the data.}
\label{fig:sameinit}
\end{figure}

\begin{figure}[ht]
\centering
\begin{subfigure}{.32\textwidth}
\centering
  \includegraphics[width=1\linewidth]{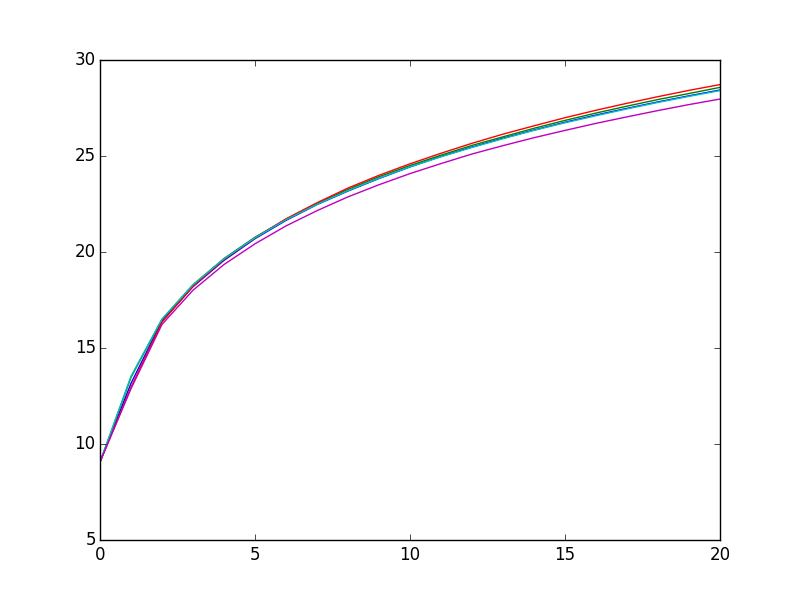}
\end{subfigure}
\begin{subfigure}{.32\textwidth}
  \centering
  \includegraphics[width=1\linewidth]{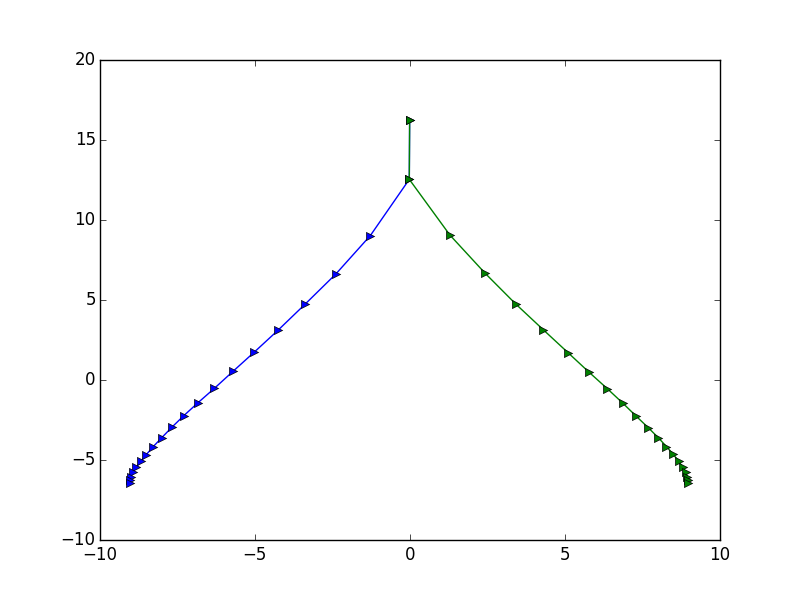}
\end{subfigure}
\begin{subfigure}{.32\textwidth}
  \centering
  \includegraphics[width=1\linewidth]{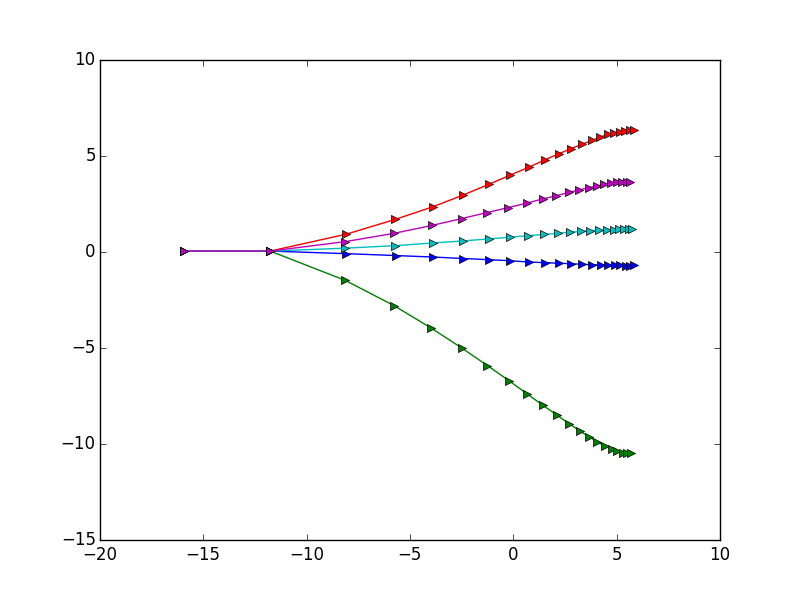}
\end{subfigure}
\caption{(a): Euclidean distances from origin after every $10$ epochs. All models hit $0.000$ error by epoch 100. All movement afterwards is through a flat region of weight space. (b) \& (c): Paths through weight-space, after 10 epochs of training followed by cloning and reshuffling.}
  \label{fig:pretrain}
\end{figure}

When we train models from the same initialization, seen in \autoref{fig:sameinit}, they nevertheless diverge, 
finding solutions far apart as measured by euclidean distance.
Interestingly, all pairs of solutions were equally far apart from each other and equally far from the origin, suggesting strong symmetry in weight space.
These observations hold even when we first pretrain the network for 10 epochs (achieving training set error around $1\%$) before cloning and shuffling (\autoref{fig:pretrain}).

\section{Conclusion}
In these experiments, we present several novel observations 
about the error surfaces of neural networks.
We showed that paths through weight space are highly nonlinear, and that local minima (albeit good ones) are abundant. 
Further, we showed that even after symmetry is broken by random initialization, 
the error surfaces of neural networks appears to be highly non-convex. 
The stochasticity introduced by reshuffling data appears to be enough to diverge otherwise identical (and partially trained) networks towards different local minima. 
Further, we showed that regions of weight space with near $0$ loss are not critical points, but large flat basins through which weights continue to travel as training continues. 

While these results are interesting, 
this work should be expanded in future iterations 
to verify that these observations hold on larger datasets
and other common neural network topologies, 
including multilayer perceptrons and LSTM recurrent neural networks.
Further, this work should be evaluated on datasets
where it is impossible for a net trained by gradient descent to achieve arbitrarily low loss.
It's possible that on such datasets, when the net is not so badly over-complete the weights truly do arrive at some critical point.

\newpage
\subsubsection*{Acknowledgments}
Zachary C. Lipton is supported by the Division of Biomedical Informatics at the University of California, San Diego, via training grant (T15LM011271) from the NIH/NLM.
Thanks to NVIDIA Corporation for their generous hardware donations.
Thanks to John Berkowitz, Anima Anandkumar, Charles Elkan, and Julian McAuley.

\bibliography{stuck}
\bibliographystyle{iclr2016_workshop}

\end{document}